\documentclass{article}

    \PassOptionsToPackage{numbers, compress}{natbib}

   \usepackage[preprint]{neurips_2020}

\usepackage[utf8]{inputenc} 
\usepackage[T1]{fontenc}    
\usepackage{hyperref}       
\usepackage{url}            
\usepackage{tabularx,booktabs}  
\usepackage{amsfonts}       
\usepackage{nicefrac}       
\usepackage{microtype}      
\usepackage{color}
\usepackage{tikz}
\usepackage{graphicx}
\usepackage{subcaption}
\usepackage{pifont}
\usepackage{caption} 
\usepackage{float}
\usepackage{wrapfig}

\newcolumntype{Y}{>{\centering\arraybackslash}X}   

\newcommand{\cmark}{\ding{51}}%
\newcommand{\xmark}{\ding{55}}%

\usetikzlibrary{shapes,decorations,arrows,calc,arrows.meta,fit,positioning}
\tikzset{
    -Latex,auto,node distance =1 cm and 1 cm,semithick,
    state/.style ={ellipse, draw, minimum width = 0.7 cm},
    point/.style = {circle, draw, inner sep=0.04cm,fill,node contents={}},
    bidirected/.style={Latex-Latex,dashed},
    el/.style = {inner sep=2pt, align=left, sloped}
}

\definecolor{orange}{gray}{0.9}

\title{Shortcomings of Counterfactual Fairness and a Proposed Modification}

\author{%
  Fabian Beigang \\
  Department of Philosophy, Logic and Scientific Method\\
  London School of Economics and Political Science\\
  \texttt{f.beigang@lse.ac.uk} \\
}

\newtheorem{definition}{Definition}[section]

\begin{document}

\maketitle

\begin{abstract}
  In this paper, I argue that counterfactual fairness does not constitute a necessary condition for an algorithm to be fair, and subsequently suggest how the constraint can be modified in order to remedy this shortcoming. To this end, I discuss a hypothetical scenario in which counterfactual fairness and an intuitive judgment of fairness come apart. Then, I turn to the question how the concept of discrimination can be explicated in order to examine the shortcomings of counterfactual fairness as a necessary condition of algorithmic fairness in more detail. I then incorporate the insights of this analysis into a novel fairness constraint, \textit{causal relevance fairness}, which is a modification of the counterfactual fairness constraint that seems to circumvent its shortcomings.
\end{abstract}

\section{Introduction}

In the contemporary debate around algorithmic fairness there is an ever increasing number of researchers who acknowledge that in order to analyze potentially discriminatory effects in algorithmic decision-making, causal relations have to be taken into account \cite{kusner,kilibertus,russell,dedeo,bonchi}. One promising approach to analyzing algorithmic fairness using causal concepts is \textit{counterfactual fairness} \cite{kusner}. Counterfactual fairness is a formal constraint on machine learning algorithms that makes use of Pearl's framework of causal modelling \cite{pearl}. It formalizes the notion that the outcome of a fair algorithmic decision-making procedure for an individual with a given protected attribute would have been the same if this protected attribute had been different. Put differently, in a fair algorithmic decision the protected attribute has no causal effect on the outcome.

In this paper, we demonstrate that counterfactual fairness does not constitute a necessary condition for an algorithm to be fair, and we subsequently propose a way how the constraint can be modified in order to remedy this problem. In section 2 we introduce the formal definition of counterfactual fairness and show by means of a hypothetical scenario that the constraint fails to provide a necessary condition for algorithmic fairness. In section 3, we turn to the question how the concept of discrimination can be explicated in order to examine the shortcomings of counterfactual fairness more systematically. In section 4, we incorporate the insights of the previous section into a novel fairness constraint, \textit{causal relevance fairness}, which is a modification of the counterfactual fairness constraint. Section 5 provides a brief summary and discussion of the project. 

The two central contributions of this work are, first, to provide a precise conceptualization of the normative notion of discrimination in the context of algorithmic decision-making that allows to analyze the adequacy of different formal fairness constraints, and second, to propose and justify a new fairness criterion that is more closely in line with the philosophical and legal notion of discrimination, and which consequently allows to evaluate an algorithm's potential to deepen existing societal injustices in a more precise way.  

\textbf{Related work.} A number of approaches have been developed which allow for some effect of the protected attribute on the predictor. \cite{chiappa} and \cite{loftus} provide a frameworks for analyzing whether individual causal paths in a model satisfy counterfactual fairness, allowing for the possibility of some of those links to not be subject to fairness constraints. \cite{kilibertus} presents a notion of fairness in which causal paths from protected attribute to predictor that are not mediated by proxy variables are considered fair. Practical limitations of counterfactual fairness have been addressed in \cite{kilibertus2} and \cite{unid}.

\section{Counterfactual fairness and its non-necessity}

Counterfactual fairness builds on the idea that discrimination means differential treatment of two individuals on the grounds that one of them has a certain sensitive trait (the so-called \textit{protected attribute}), such as a particular religion, ethnicity, or gender. Accordingly, a non-discriminatory and hence fair algorithmic decision is one in which the protected attribute does not make a difference - the protected attribute is not a cause of the outcome. This is formalized as the requirement that an individual with a given value of a protected attribute would have received the same outcome that it actually received, had this protected attribute taken a different value while everything else that is not causally downstream of the protected attribute had stayed the same.

In order to make this notion mathematically precise, a number of definitions from Pearl's \cite{pearl} framework of causal modelling are introduced. First, a \textit{causal model} is defined as a triple $(U,V,F)$ such that (i) $U$ is a set of variables determined by factors \textit{not} in the model, (ii) $V$ is a set $\{V_1, V_2, ..., V_n\}$ of variables determined by other variables in the model, and (iii) $F$ is a set of functions $\{f_1, f_2, ..., f_n\}$ such that each function $f_i$ is a mapping from the respective domains of $U \cup (V \setminus V_i)$ to the domain of $V_i$ \citep[p. 203]{pearl}.\footnote{Note that, following \citep[p. 9]{pearl}, we use the same notation for variables and sets of variables, assuming that the latter define compound variables the domains of which are the Cartesian products of the domains of the individual variables in the respective sets.}


Next, a \textit{submodel} of a causal model $M$ is a causal model $M_x = (U, V, F_x)$ where $F_x = \{ f_i : V_i \notin X \} \cup \{X = x\}$ for a particular realization $X = x$ of a set of variables $X \subseteq V$. Using the notion of a submodel, Pearl defines \textit{counterfactuals} of the form "The value that Y would have obtained, had X been x" (for $X,Y \subseteq V$) as denoting the \textit{potential response} $Y_x(u)$, where potential response means the solution for $Y$ of the set of equations $F_x$ in the submodel $M_x$ of $M$ \citep[p. 204]{pearl}.

Given a causal model $M$ and a probability distribution $P(u)$ over the domain of $U$, the conditional probability of a counterfactual "If it were $X = x$, then $Y = y$" given evidence $e$ can be evaluated by (1) updating $P(u)$ by conditioning on evidence e in order to obtain $P(u|e)$, (2) generate the submodel $M_x$ of $M$ obtained by setting $X = x$, (3) use the submodel $M_x$ and conditional probability distribution $P(u|e)$ to compute the probability of $Y=y$.

This allows to define counterfactual fairness in a mathematically precise way. Let the variable $\hat{Y}$ denote the \textit{predictor} of the algorithmic decision system, $Y$ the outcome to be predicted (or, the \textit{target variable}), $A$ the set of \textit{protected attributes}, $X$ the (remaining) \textit{input features}, and $U$ the set of relevant \textit{latent variables}.  

\begin{definition}[Counterfactual fairness]
The predictor $\hat{Y}$ is counterfactually fair if under any context $X = x$ and $A = a$, $P(\hat{Y}_a (u) = y | X = x, A = a) = P(\hat{Y}_{a'} (u) = y | X = x, A = a)$ for all $y$ and any value $a'$ attainable by $A$
\end{definition}

Counterfactual fairness gives the intuitively right verdict in a number of hypothetical and actual cases. However, the definition does have its shortcomings. A formal fairness constraint should provide necessary and sufficient conditions for  fair algorithmic decisions. In other words, it should be the case that whenever we intuitively judge an algorithmic decision to be fair, the formal fairness constraint is satisfied, and whenever we judge a decision to be unfair, the constraint is violated. Consequently, we can show that a formal fairness constraint is not a sufficient condition for algorithmic fairness if we can show that there are situations in which the constraint is satisfied, yet the algorithmic decision is intuitively unfair. Analogously, we can show that a formal fairness constraint is not a necessary condition for algorithmic fairness if we can show that there are situations in which the constraint is not satisfied despite the algorithmic decision being intuitively fair. Now, consider the following scenario:

\textbf{Scenario: Predicting accident rates.} A government's department for transport applies a predictive model to predict individuals' accident rates $Y$ in order to determine whether to issue a driving licence. To this end, the department measures each individual's driving ability $X$ on the basis of which the model produces a prediction $\hat{Y}$ of their estimated accident rates. By assumption, $X$ is influenced by whether an individual has a (severe) visual impairment $A$, as well as by further variables (which are not explicitly modelled in the graph).

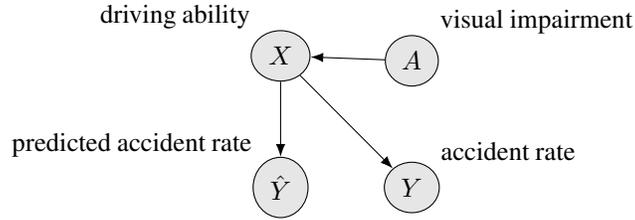
\begin{figure}
    \centering

        \begin{tikzpicture}
    \node[state, fill=orange,  label=north west:predicted accident rate] (x) at (0,0) {$\hat{Y}$};

    \node[state, fill=orange, label=north east:accident rate] (y) [right =of x] {$Y$};
    \node[state, fill=orange,  label=north west:driving ability] (z) [above =of x] {$X$};
    \node[state, fill=orange,  label=north east:visual impairment, align=center] (a) [above =of y] {$A$};

    \path (z) edge   (y);
    \path (z) edge  (x);
    \path (a) edge (z);

\end{tikzpicture}
   \caption{Causal structure of "Predicting accident rates"}
    \label{fig:my_label}
\end{figure}

The causal structure of this scenario is depicted in figure 1. On an intuitive level it seems fair that predictions of a person's accident rate are (at least partly) based on their ability to drive a car, even though the latter is influenced by the fact that an individual has a severe visual impairment, which, given its status as disability, constitutes a protected attribute. Consequently, the prediction would violate counterfactual fairness due to $A$'s (mediated) causal influence on $\hat{Y}$. In a counterfactual world in which a person who is actually visually impaired had no such impairment, the prediction would not be the same. Given the above analysis is accepted, we can conclude that counterfactual fairness is not a necessary condition for algorithmic fairness.


\section{An explication of the concept of wrongful discrimination}

In order to understand what is missing in the definition of counterfactual fairness, we first need to get a grasp on the concept of discrimination. This will allow to see more clearly which notions from moral discourse the definition of counterfactual fairness corresponds to. We will here follow Eidelson's \cite{eidelson} treatment of the subject.

In Eidelson's view, discrimination occurs when a person is treated differently than another in a certain dimension, where this differential treatment constitutes a comparative disadvantage for the person, and occurs on the grounds of a perceived difference between the discriminatee and another (actual or hypothetical) person. Hence, discrimination - understood as a non-normative concept - consists of two components: the \textit{differential treatment}, and the \textit{explanatory relation} between a perceived difference and this differential treatment.  

The normative question about the concept of discrimination is under which conditions discrimination is wrong \cite{moreau,alexander}. Clearly, discrimination as defined above is not always morally wrong: denying a visually disabled person a driving licence because of their disability constitutes discrimination in the above sense, but not a morally wrong act. According to Eidelson, what distinguishes wrongful discrimination from permissible discrimination is that it \textit{fails to respect the discriminatee's standing as a person}. This, in turn, involves two aspects: (1) recognizing that a person's moral worth is equal to all others, and (2) treating a person as an individual. We hence arrive at a definition of \textit{wrongful discrimination} against an individual $i_1$ which involves the conjunction of the following three conditions:

\begin{itemize}
    \item \textbf{(Differential Treatment Condition)} Individual $i_1$ is treated less favourably in respect of some dimension $W$ than some actual or counterfactual other, individual $i_2$
    \item \textbf{(Explanatory Condition)} A (perceived) difference between $i_1$ and $i_2$ with regards to the sensitive trait $A$ figures in the explanation of this differential treatment
    \item \textbf{(Wrongfulness Condition)} This differential treatment on the basis of A constitutes a (1) failure to recognize that $i_1$'s moral worth is equal to $i_2$'s, or (2) a failure to treat $i_1$ as an individual.
\end{itemize}

In order to apply this definition to algorithmic systems, we need to examine how the third condition can hold in automated decision-making systems. As it is unclear what it would mean that an automated system fails to recognize a person's moral worth, we will focus on how it could fail to treat a person as an individual. According to Eidelson \citep[p. 142]{eidelson}, one way in which this can happen is when decisions are based on wrongful generalizations. A generalization can be wrongful when in the process of arriving at a prediction, it was not given "reasonable weight to evidence about the ways [a person] has exercised her autonomy in giving shape to her life, where this evidence is reasonably available and relevant to the determination at hand" \citep[p. 144]{eidelson}, or when the prediction is "made in a way that disparages [the person's] capacity to make [...] choices as an autonomous agent" \citep[p. 144]{eidelson}  in the case that predictions are concerned with the person's choices. 

To illustrate this, imagine a landlord who judges on the basis of an applicant's religion that the applicant will not pay their rent reliably, despite the fact that the applicant provides evidence of a secure job and references from previous landlords. The landlord generalizes from the supposition that people that have the applicant's religion do not pay their rent reliably. In doing so, the landlord clearly fails to acknowledge evidence about how the applicant has autonomously shaped their life, for instance in terms of what kind of career to pursue, which evidences reliability. Furthermore, the landlord ignores evidence of the choices the applicant made in the past, in particular with regard to paying the rent reliably, which was readily available in the form of references from previous landlords. Therefore, this act is an instance of a decision that is based on a wrongful generalization.

Wrongful generalizations are thus generalizations in which weight is given to evidence in an inadequate way. In particular, not enough weight is given to evidence about a person's character traits, where these character traits are relevant to the property that is to be predicted (e.g. reliability), or not enough weight is given to the person's choices, where these choices are relevant to the predicted property (e.g. criminal behaviour). Instead, too much weight is given to a person's membership in a demographic group (i.e. a protected attribute), despite this not being directly (or only to a lesser extent) relevant to the predicted property. 

Consequently, treating someone as an individual means giving the right weight to all the relevant factors in making a judgment about a person. This entails that a person's protected attributes can only influence the prediction to the degree to which they are actually relevant. If it influences the prediction to a higher degree, then this means that not enough weight is attributed to the person's relevant character traits or the person's relevant choices as an autonomous agent.

The above explication of wrongful discrimination allows for a causal interpretation that is applicable to algorithmic predictions. To this end, two core concepts have to be defined in causal terms: a property's \textit{relevance} to a prediction, and a property's \textit{influence} on a prediction. The former can be defined as the actual causal effect of a property on the predictor, the latter as the actual causal effect of a property on the variable that is to be predicted. The definition of wrongful discrimination can then be reformulated as stating that wrongful algorithmic discrimination contains a (i) difference between two individuals $i_1$ and $i_2$'s (conditional) probability distribution over the predictor $\hat{Y}$, (ii) where this difference is due to an influence of protected attribute $A$ on the predictor $\hat{Y}$, and (iii) the influence of the protected attribute $A$ on the predictor exceeds $A$'s (causal) relevance to the prediction. 

Following Kusner et al. \cite{kusner} we define the \textit{actual causal effect}\footnote{Kusner et al. call actual causal effects "counterfactual effects"}  $\alpha_{Y|a}$ of $A = a$ on Y under context $X = x$ as $\alpha_{Y|a} = P(Y_a (u) | A = a, X = x) - P(Y_{a'} (u) | A = a, X = x)$, where $A = a'$ is the Boolean negation of $A = a$. With this definition at hand, we can now provide a full definition of \textit{wrongfully discriminatory algorithmic predictions} (against $i_1$ as compared to the actual or hypothetical individual $i_2$, where $A = a$ denotes the protected attribute) as involving the following three conditions:

\begin{itemize}
    \item \textbf{(Differential Treatment Condition)} $P(\hat{Y}^{(1)} | W) \neq P(\hat{Y}^{(2)} | W)$, for some $W$
    \item \textbf{(Explanatory Condition)} $\alpha_{\hat{Y}|a} > 0$
    \item \textbf{(Wrongfulness Condition)} $\alpha_{Y|a} < \alpha_{\hat{Y}|a}$
\end{itemize}

This causal interpretation allows to see that counterfactual fairness is a constraint that is violated whenever the first two conditions are satisfied - namely, in cases of differential predictions that are explained by an individual's protected attributes. This means, whenever counterfactual fairness is violated, this is a case of discrimination in the non-normative sense. However, counterfactual fairness fails to distinguish cases of wrongful discrimination from permissible cases of discrimination. We conclude that counterfactual fairness is too strong a definition of fairness: it not only ensures the absence of wrongful discrimination, but the absence of discrimination (in the non-moralized sense) whatsoever. This, as we have seen above, can itself generate unfair decisions. In the next section we will modify counterfactual fairness so that it also takes the \textit{wrongfulness condition} into account.  

\section{Causal relevance fairness}

An algorithmic prediction is unfair according to the above analysis if the presence of a protected attribute ($A = a$) has a stronger causal effect on a prediction of a variable ($\hat{Y}$) than on the predicted variable ($Y$) itself. In other words, when the protected attribute’s influence on a prediction exceeds its causal relevance. We will call the constraint that ensures the absence of this type of unfairness \textit{causal relevance fairness}. It can be formalized as follows:

\begin{definition}[Causal relevance fairness] The predictor $\hat{Y}$ satisfies causal relevance fairness if $\alpha_{\hat{Y}|a} \leq \alpha_{Y|a}$ for any protected attribute $A = a$ and context $X=x$
\end{definition}

It is easy to see that counterfactual fairness is a limiting case of causal relevance fairness, namely in exactly those cases when the protected attribute is not causally relevant, i.e. $\alpha_{Y|a} = 0$, since this requires that the protected attribute does not have an influence on the predictor, i.e.  $\alpha_{\hat{Y}|a} = 0$, and this, by our definition of causal effect, is equivalent to $P(Y_a (u) | A = a, X = x) - P(Y_{a'} (u) | A = a, X = x) = 0$. Hence, whenever the protected attribute is causally irrelevant to the predicted target variable, causal relevance fairness coincides with counterfactual fairness. Since in most of the striking cases of discrimination, the protected attribute is in fact causally irrelevant for the predicted variable - think of using race for predicting recidivism \cite{ulmer,aliprantis}, or gender for predicting quantitative reasoning capabilities \cite{else} - counterfactual fairness gives the right verdict in many cases. This explains the intuitive appeal of counterfactual fairness. 

\begin{figure} [t]
\centering
\begin{tabular}{ccccc}
\scalebox{0.9}{
\begin{tikzpicture}
    \node[state, fill=orange] (x) at (0,0) {$\hat{Y}$};

    \node[state, fill=orange] (y) [right =of x] {$Y$};
    \node[state, fill=orange] (z) [above =of x] {$X$};
    \node[state, fill=orange] (a) [above =of y] {$A$};

    \path (z) edge (y);
    \path (z) edge (x);
    \path (a) edge[thick] (x);

\end{tikzpicture}} &
\scalebox{0.9}{
\begin{tikzpicture}
    \node[state, fill=orange] (x) at (0,0) {$\hat{Y}$};

    \node[state, fill=orange] (y) [right =of x] {$Y$};
    \node[state, fill=orange] (z) [above =of x] {$X$};
    \node[state, fill=orange] (a) [above =of y] {$A$};

     \path (z) edge (y);
    \path (z) edge (x);
    \path (a) edge[thick] node[right] {$\beta$} (y);
    \path (a) edge[thick] node[above] {$\alpha$} (x);

\end{tikzpicture}
} &
\scalebox{0.9}{
\begin{tikzpicture}
    \node[state, fill=orange] (x) at (0,0) {$\hat{Y}$};

    \node[state, fill=orange] (y) [right =of x] {$Y$};
    \node[state, fill=orange] (z) [above =of x] {$X$};
    \node[state, fill=orange] (a) [above =of y] {$A$};

     \path (z) edge (y);
    \path (z) edge (x);
    \path (a) edge[thick] node[right] {$\gamma$} (y);
    \path (a) edge[thick] node[above] {$\alpha$} (x);

\end{tikzpicture} 
}
&
\scalebox{0.9}{
\begin{tikzpicture}
    \node[state, fill=orange] (x) at (0,0) {$\hat{Y}$};

    \node[state, fill=orange] (y) [right =of x] {$Y$};
    \node[state, fill=orange] (z) [above =of x] {$X$};
    \node[state, fill=orange] (a) [above =of y] {$A$};

    \path (z) edge (y);
    \path (z) edge (x);
    \path (a) edge[thick] (y);

\end{tikzpicture}
}
&
\scalebox{0.9}{
      \begin{tikzpicture}
    \node[state, fill=orange] (x) at (0,0) {$\hat{Y}$};

    \node[state, fill=orange] (y) [right =of x] {$Y$};
    \node[state, fill=orange] (z) [above =of x] {$X$};
    \node[state, fill=orange] (a) [above =of y] {$A$};

    \path (z) edge (y);
    \path (z) edge (x);

\end{tikzpicture}
}\\
\textbf{(a)}  & \textbf{(b)} & \textbf{(c)} & \textbf{(b)} & \textbf{(c)} \\[6pt]
\end{tabular}
\caption{ Five different normatively relevant causal structures.}
\label{fig:Name}
\end{figure}
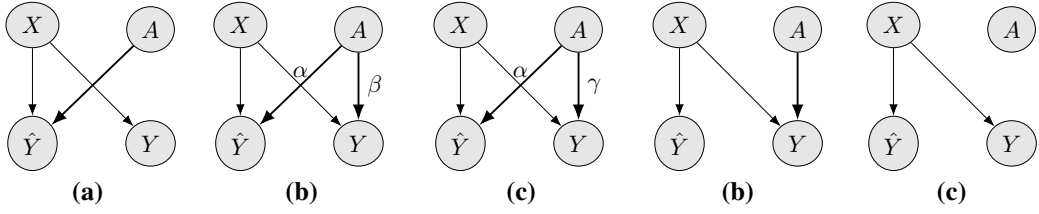

\begin{table}[ht]
\centering
\caption{Comparative evaluation of different causal structures.}

\begin{tabularx}{\textwidth}{l *{5}{Y}}

\toprule

 & \multicolumn{5}{c}{\textbf{Causal structure}}  \\
 \cmidrule(l){2-6} 
 \textbf{Criterion} & \textbf{(a)} & \textbf{(b)} [$\alpha > \beta$] & \textbf{(c)} [$\alpha \leq \gamma$] & \textbf{(e)} & \textbf{(d)} \\
\midrule
 Counterfactual fairness  & \xmark &  \xmark &  \xmark & \cmark & \cmark \\
 Causal relevance fairness  & \xmark & \xmark & \cmark & \cmark & \cmark \\
\bottomrule
\end{tabularx}
\label{my_label}
\end{table}

Let us now turn to the discussion of different causal structures and their normative evaluation using the causal relevance fairness criterion. Figure 2(a) depicts a causal graph in which the protected attribute $A$ has a causal effect on the predictor $\hat{Y}$, but no causal effect on the predicted variable $Y$. This is a typical structure of a wrongfully discriminatory prediction. For example, target variable $Y$ could stand for the recidivism rate, $\hat{Y}$ for predicted recidivism, and $A$ for ethnicity. As is well known, disparities in crime rates between different ethnicities vanish when certain socioeconomic factors are controlled for \cite{ulmer, aliprantis}, showing that ethnicity is not a cause of criminal behavior. If a prediction about whether an individual will recidivate is made on the basis of ethnicity, this is hence discriminatory. Accordingly, causal relevance fairness and counterfactual fairness would be violated. 

Figure 2(b) shows a causal structure in which A does have an effect on $\hat{Y}$ and $Y$, but, by assumption, the effect $\alpha$ on the former is greater than the effect $\beta$ on the latter. An example of this would be a situation in which a prediction is to be made about the performance of a student with a learning disability, in which it is predicted that the student will perform terribly on the basis of their learning disability, while in reality the learning disability only has a slight effect on the student's performance. This constitutes a case of a wrongfully discriminatory prediction. Both, counterfactual fairness and causal relevance fairness are, as we would expect, violated.

Figure 2(c) is the causal structure in which counterfactual fairness and causal relevance fairness come apart. Here, $A$ has a causal effect on both, the predictor $\hat{Y}$ and the target variable $Y$. Assuming that the causal effect $\alpha$ that $A$ has on $\hat{Y}$ is equal or lower than the causal effect $\gamma$ it has on $Y$, this causal structure is one of permissible discrimination. For example, target variable $Y$ could stand for the risk of being involved in an accident, $\hat{Y}$ for the prediction thereof, and the protected attribute $A$ could stand for a visual disability. As a visual disability might arguably have a causal effect on the risk of being involved in an accident, it is permissible that the visual disability influences the prediction of the risk of being involved in an accident, given that the latter does not exceed the former. This is the structure of the scenario presented in section 2. Causal relevance fairness would be satisfied in this causal structure, whereas counterfactual fairness would not. 

In figure 2(d), a causal structure is shown in which the protected attribute $A$ has an effect on the target variable $Y$, but not on the predictor $\hat{Y}$. This is a case of ignoring a potentially relevant protected attribute in making the prediction. In many cases, this seems permissible, as, for instance, in cases of affirmative action. Both, counterfactual fairness and causal relevance fairness are satisfied in this causal structure. 


Lastly, in figure 2(e), the causal structure does neither involve a causal link from the protected attribute $A$ to the target variable $Y$, nor to the predictor $\hat{Y}$. This is what could be considered a standard case of fair prediction. Imagine, say, that a bank intends to predict someone's creditworthiness $Y$ (defined as the probability of the applicant paying back a loan). The protected attribute could be ethnicity $A$. Not taking ethnicity into account in predicting creditworthiness is fair (and even morally required) because ethnicity has no effect on someone's creditworthiness. Both, causal relevance fairness and counterfactual fairness are satisfied in cases with this causal structure. 

Table 1 summarizes the evaluation of the four different causal structures in light of the causal relevance fairness criterion and the counterfactual fairness criterion, respectively. There is no difference in the normative evaluation of causal structures (a), (b), (d), and (e). Only the causal structure in (c) is evaluated differently. Assuming that a majority of people would agree with our intuition that the structure in (c) is morally permissible, it seems that casual relevance fairness captures intuitions about fairness more closely than counterfactual fairness, while maintaining all of the latter's advantages over other non-causal fairness criteria.



\section{Conclusion}

I presented a criticism of one of the most widely discussed causal definitions of algorithmic fairness, counterfactual fairness, by demonstrating that it does not provide a necessary condition for fair algorithmic decision-making. After analyzing the concept of discrimination, I proposed a modification of the criterion which aligns the mathematical fairness constraint more closely with the philosophical and legal notion of discrimination. The discussion of different possible causal structures showed that the modified constraint can solve the problem of non-necessity, while maintaining all the advantages of counterfactual fairness.






\end{document}